\pdfoutput=1
\PassOptionsToPackage{prologue,dvipsnames}{xcolor}

\documentclass[11pt]{article}

\usepackage[]{acl}

\usepackage{times}
\usepackage{latexsym}

\usepackage{multirow}
\usepackage{graphicx}
\usepackage{hyperref}
\usepackage{xcolor}

\usepackage{tikz}
\newcommand{\tikzxmark}{%
\tikz[scale=0.23] {
    \draw[line width=0.7,line cap=round] (0,0) to [bend left=6] (1,1);
    \draw[line width=0.7,line cap=round] (0.2,0.95) to [bend right=3] (0.8,0.05);
}}
\newcommand{\tikzcmark}{%
\tikz[scale=0.23] {
    \draw[line width=0.7,line cap=round] (0.25,0) to [bend left=10] (1,1);
    \draw[line width=0.8,line cap=round] (0,0.35) to [bend right=1] (0.23,0);
}}

\usepackage{arydshln}
\usepackage{graphicx}

\usepackage[T1]{fontenc}

\usepackage[utf8]{inputenc}

\usepackage{microtype}

\usepackage{inconsolata}

\title{Comparing Knowledge Sources for Open-Domain \\Scientific Claim Verification}

\author{Juraj Vladika \and Florian Matthes \\
  Department of Computer Science \\ Technical University of Munich \\ Garching, Germany \\
  \texttt{ \{juraj.vladika, matthes\}@tum.de} \\}

\begin{document}
\maketitle
\begin{abstract}

The increasing rate at which scientific knowledge is discovered and health claims shared online has highlighted the importance of developing efficient fact-checking systems for scientific claims. The usual setting for this task in the literature assumes that the documents containing the evidence for claims are already provided and annotated or contained in a limited corpus. This renders the systems unrealistic for real-world settings where knowledge sources with potentially millions of documents need to be queried to find relevant evidence. In this paper, we perform an array of experiments to test the performance of open-domain claim verification systems. We test the final verdict prediction of systems on four datasets of biomedical and health claims in different settings. While keeping the pipeline's evidence selection and verdict prediction parts constant, document retrieval is performed over three common knowledge sources (PubMed, Wikipedia, Google) and using two different information retrieval techniques. We show that PubMed works better with specialized biomedical claims, while Wikipedia is more suited for everyday health concerns. Likewise, BM25 excels in retrieval precision, while semantic search in recall of relevant evidence. We discuss the results, outline frequent retrieval patterns and challenges, and provide promising future directions.
\end{abstract}

\section{Introduction}

The fast promulgation of knowledge in the digital world has made keeping track of information trustworthiness a challenging endeavor. 
In particular, science and health have become popular talking points and brought with it an abundance of medical advice that permeate online resources \cite{swire2020public}.
A report by the Pew Research Center \cite{fox2013health} found that over one-third of American adults have searched the Internet for medical conditions and asked it medical questions before going to a medical professional.  The sought information ranged from self-diagnosis to finding medications. 

\begin{figure}[t]
  \centering
  \includegraphics[width=0.77\linewidth]{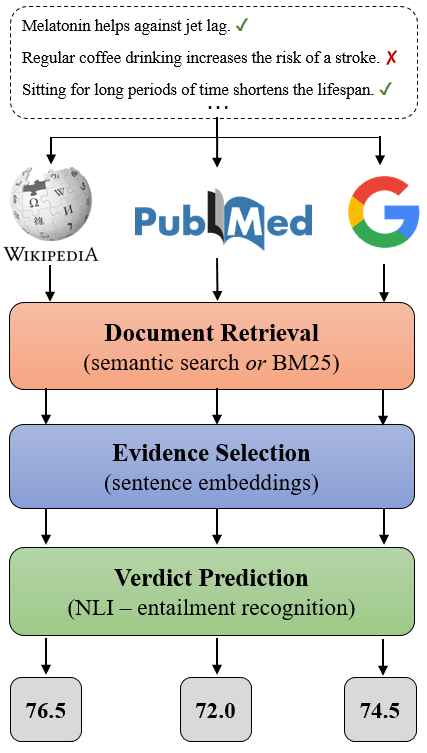}
  \caption{The experimental setup of the study: scientific claims are passed through a fixed pipeline using three different knowledge sources, resulting in different final prediction performance (as measured by F1 score).}
  \label{fig:pipeline}
\end{figure}

Automated solutions for claim verification based on Natural Language Processing (NLP) have emerged as a potential aid to help with bringing light into the information overload \cite{Nakov2021AutomatedFF}. While most work in the automated fact-checking domain is concerned with claims related to politics, society, rumors, and general misinformation, there has been an increasing interest in fact-checking of scientific and biomedical claims \cite{kotonya-toni-2020-explainable-automated, wright-etal-2022-generating}. The task of automated claim verification consists of retrieving evidence for a claim being checked and then predicting a veracity label based on the discovered evidence. The most common setting for this task either already provides the source document that will contain evidence for the claim or works over a limited, manually constructed collection of documents \cite{saakyan-etal-2021-covid}. While this is an important step in developing models capable of reading comprehension and detecting which spans provide evidence in a given context, this is not a realistic setting for automated claim verification systems deployed in the real world. In such a scenario, the documents containing evidence are not known and knowledge bases containing them can possibly contain millions of documents. Moreover, with the rise of medical assistants and conversational agents in healthcare, many users are turning to these systems as a source of health-related information and medical support \cite{valizadeh2022ai}.


To address these research gaps, we perform an array of experiments that test the performance of NLP systems for claim verification in the open domain. In the experiments, we keep the parts of the verification pipeline concerned with evidence sentence selection and verdict prediction fixed, and vary the knowledge source being used and information retrieval techniques being deployed to query the databases. Since the final goal of fact-verification systems is to provide a verdict on the correctness of a claim, we measure the usefulness of knowledge sources and retrieval techniques by looking at verdict prediction scores. For this purpose, we leverage four English datasets of biomedical and health claims that contain gold annotations stemming from domain experts. We use the veracity labels of claims in datasets as ground truth.

We opt for three large-scale knowledge sources: PubMed, the cardinal collection of biomedical research publications; Wikipedia, as the largest publicly curated encyclopedia of human knowledge; and Google search results (representing "the whole web"), which is a straightforward and intuitive way how users seek information. 
Finally, we perform a qualitative analysis of retrieved evidence for some interesting example claims, present the insights from results, and provide future directions.

Our contributions are as follows:

\begin{enumerate}
  \item We test the claim verdict prediction performance of a fixed fact-verification system on four biomedical fact-checking datasets by using three different knowledge sources (PubMed, Wikipedia, Google Search).
  \item We compare the final label prediction performance by retrieving evidence using different techniques (sparse retrieval with BM25 and semantic search with dense vectors).
  \item We provide a qualitative error analysis of retrieved evidence for different types of claims and provide insights and future directions for open-domain claim verification.
\end{enumerate}

We make the data and code of the experiments available in a GitHub repository.\footnote{\url{https://www.github.com/jvladika/comparing-knowledge-sources}}

\section{Foundations}

\subsection{Pipeline for Automated Claim Verification}
The systems for automated claim verification and automated fact-checking are usually modeled as a framework with three components, where each component is a well-established NLP task \cite{zeng2021automated}. This framework is a three-component pipeline (Figure \ref{fig:pipeline}) consisting of (1) document retrieval; (2) evidence selection; (3) verdict prediction. We are mostly concerned with how the document retrieval part affects the further entailment process. That is why we fix the evidence-sentence selection model and the entailment prediction model. This way, the quality of the data source and the retrieval technique are the most important variables being tested. Two of these subtasks, or even all three \cite{ARSJoint}, can be learned together in a joint system with a shared representation. For the sake of simplicity of testing, we choose a pipeline system that performs each task sequentially. 

In document retrieval, given a claim $c$ and a corpus of $n$ documents $D = \{d_1, d_2, ..., d_n\}$, the task is to select top $k$ most relevant documents $g_1, ..., g_k$ for the claim (query), with a function $w(c, d)$. After the documents are retrieved, the next step is to select evidence sentences that serve as rationale in making a decision regarding the claim's veracity. From $m$ candidate sentences $s_1, s_2, ..., s_m$ comprising the selected documents, top $j$ sentences are selected as evidence sentences $\vec{e} = \{e_1, e_2, ..., e_n\}$ with a function $z(c, s)$. Finally, a verdict prediction function is trained to predict $y(c, \vec{e}) \in \{\textsc{Supported}, \textsc{Refuted}\}$. This means the final task is essentially a binary classification task with two classes, which makes it suitable for evaluation with standard classification metrics: precision, recall, and binary F1. 

Since we are focusing on testing the influence of the knowledge source on the final claim verdict prediction, we experiment with different knowledge sources $D$ and retrieval functions $w(c, d)$. Other components of the pipeline are fixed to make a fair comparison. After testing the values of $k$ and $j$ with different values in the set of $\{1, 3, 5, 10, 20\}$, we set both to be $10$ since it provided the best F1 performance and the best trade-off between covering enough content while not cluttering with too much noise. This means we retrieve the top 10 documents and then select the top 10 sentences from them. For $z(c, s)$, we select the model \textsc{Spiced} \cite{wright-etal-2022-modeling}, which is a sentence similarity model that catches paraphrases of scientific claims well and recently set state-of-the-art performance in evidence selection on a couple of scientific claim-verification datasets. For the verdict predictor $y(c, \vec{e})$, we first considered specialized biomedical language models \cite{vladika2023diversifying}. In the end, we chose the DeBERTa-v3 model \cite{he2021deberta}, since it had the best performance and was shown to be an exceptionally powerful model for textual entailment recognition on the GLUE benchmark. We use a version of DeBERTa-v3 additionally fine-tuned on various Natural Language Inference (NLI) datasets.\footnote{\url{https://huggingface.co/MoritzLaurer/DeBERTa-v3-large-mnli-fever-anli-ling-wanli}} It should be noted that we use these two models out-of-the-box and do not fine-tune them on any of our datasets in any experiment. This is an intentional zero-shot setting that aims to verify the real-world situation of using a system on yet unknown claims.


\subsection{Datasets}
We choose four English datasets of biomedical and health claims, built for different purposes.

\textsc{SciFact} \cite{wadden-etal-2020-fact} is a dataset of 1,109 claims (in test and dev set) which were expert-written from citation sentences found in biomedical research publication abstracts. These publications originate from PubMed, which is one of the databases queried in our paper. 

\textsc{PubMedQA} \cite{jin-etal-2019-pubmedqa} is a dataset of 1,000 labeled claims that were generated from abstract of biomedical papers originating from PubMed. Even though more of a question-answering dataset in nature, it also provides yes/no/maybe labels which make it usable as a fact-checking dataset. 

\textsc{HealthFC} \cite{vladika2023healthfc} is a dataset of 750 claims concerning everyday health and spanning various topics like nutrition, immune system, mental health, and physical activity. The claims originate from user inquiries and they were checked by a team of medical experts using clinical trial reports and systematic reviews as the main evidence source. All the claim verdict explanations are described in a user-friendly language.

\textsc{CoVert} \cite{mohr-whrl-klinger:2022:LREC} is a dataset of 300 claims related to health and medicine, which are all causative in nature (such as "\textit{vaccines cause autism}"). All the claims originate from Twitter, which means some claims are written informally and thus make an additional challenge by providing a real-world scenario of misinformation checking.

\begin{table}[htpb]
\begin{tabular}{p{20mm}|p{19mm}|ccc}
\hline
\textbf{Dataset}  & \textbf{Domain}                & \tikzcmark & \tikzxmark & $\sum$ \\
\hline
\textsc{SciFact}  & biomedical research   & 456 & 237 & 693 \\ \hdashline
\textsc{PubMedQA} & biomedical research   & 552 & 338 & 890 \\ \hdashline
\textsc{HealthFC} & consumer health       & 202 & 125 & 327 \\ \hdashline
\textsc{CoVert}   & health misinformation & 198 & 66  & 264 \\ 
\hline
\end{tabular}

\caption{\label{tab:datasets} The four datasets used in the experiments, including their domain and label distribution.}
\end{table}

For all of the datasets, we leave out any claims labeled with \textsc{Not Enough Information (NEI)} label. This is because some datasets do not include this label, and those that do include it define it differently. For \textsc{SciFact}, NEI means no evidence documents are present in their internal corpus. For \textsc{HealthFC}, NEI means no conclusive evidence for the claim was found in any clinical trials. Table \ref{tab:datasets} shows the final distribution of labels for each dataset, after leaving out the claims with not enough information.

\section{Experiment Setup}
\subsection{Knowledge Sources}
For testing on Wikipedia, we used the latest available dump of English Wikipedia that we found, from May 20th, 2023, containing 6.6 million articles.\footnote{\url{https://dumps.wikimedia.org/enwiki/20230520/}} For PubMed, the US National Library of Medicine provides MEDLINE, a snapshot of currently available abstracts in PubMed that is updated once a year. We used the 2022 version found at their website.\footnote{\url{https://www.nlm.nih.gov/databases/download/pubmed_medline.html}} While this yields 33.4M abstracts, we pre-processed the data following \citet{gonzalez} and removed non-English papers, papers with no abstracts, and papers with unfinished abstracts, which yields 20.6M abstracts. For Google results, we used Google's publicly available Custom Search JSON API.\footnote{\url{https://developers.google.com/custom-search/v1/overview}}

\subsection{Document Retrieval Techniques}
We test the performance of two different document retrieval techniques, namely a sparse one and a dense one. Since both types of approaches are deployed in modern search systems, we want to see how much of a difference they make in finding appropriate documents that can verify a claim. As a representative sparse technique, we opt for BM25, an improvement over TF-IDF that takes into account term frequency, document length, and inverse document frequency. Despite its simplicity, it has proven to be a cornerstone of information retrieval approaches due to comparative performance to more sophisticated neural approaches \cite{kamphuis2020bm25}. 

Recently, with the advance of large language models, encoding both the claim and documents with dense vector embeddings and then searching most similar vectors with cosine similarity has proven to be a powerful retrieval method \cite{karpukhin-etal-2020-dense}. A particularly successful recent approach is SimCSE \cite{gao-etal-2021-simcse}, which uses contrastive learning and entailment-based training to enhance similarity scoring. We chose a biomedical variation BioSimCSE \cite{kanakarajan-etal-2022-biosimcse} which fits our use case. For dense retrieval, we encode the entirety of our PubMed corpus and Wikipedia corpus with BioSimCSE and store the embeddings. For sparse retrieval, we construct an inverted index out of Wikipedia and PubMed corpora and later query it using BM25 metrics. After selecting the top 10 documents in each method, the top 10 most similar sentences were taken and jointly with claim the verdict was predicted based on the entailment relation. For Google Search, we took the top 10 returned Google snippets as "evidence sentences" that we then concatenate and use as the evidence block for label prediction. All the experiments were run on a single Nvidia V100 GPU card, in a single run. One run on one dataset, one knowledge source, and one retrieval technique costed one computation hour.

\subsection{Baseline}

To establish a baseline, we first run the system with the gold evidence provided with each of the four datasets. These are the sentences or snippets that were given by the annotators or creators of the respective datasets. The performance is shown in Table \ref{tab:gold}. It should be noted that this performance is different from those reported in papers introducing these datasets because we remove the claims labeled with NEI and we also did not fine-tune the model on the datasets. This is an intentional choice because the idea is to test the systems in a zero-shot / off-the-shelf setting. We expect the results in our experiments to be lower than this because having annotated evidence is an easier setting than the open-domain claim verification where the evidence needs to be discovered.

\begin{table}[htpb]
\centering
{
\def\arraystretch{1.3}
\begin{tabular}{c|ccc}
\hline
\multicolumn{1}{l}{} & \multicolumn{3}{c}{\textbf{Gold Evidence}} \\ \hline
\textbf{Dataset} & \textbf{Precision} & \textbf{Recall} & \textbf{F1 Score}  \\ \hline
\textsc{SciFact} & $77.9$ & $88.4$ & \textbf{\color{Dandelion}{82.8}}   \\ 
\textsc{PubMedQA} & $74.4$ & $80.4$ & \textbf{\color{Dandelion}{77.3}}  \\
\textsc{HealthFC} & $80.5$ & $83.4$ &  \textbf{\color{Dandelion}{81.9}} \\ 
\textsc{CoVert} & $80.7$ & $86.4$ &  \textbf{\color{Dandelion}{83.4}} \\
 \hline
\end{tabular}

}
\caption{ \label{tab:gold} Results of final verdict prediction over four datasets using the gold evidence sentences provided with the datasets.}
\end{table}

\begin{table*}[htpb]
\centering
{
\def\arraystretch{1.3}
\begin{tabular}{c|c|ccc|ccc}
\hline
\multicolumn{2}{l}{} & \multicolumn{3}{c}{\textbf{BM25}} & \multicolumn{3}{c}{\textbf{Semantic Search}} \\ \hline
\textbf{Source} & \textbf{Dataset} & \textbf{Precision} & \textbf{Recall} & \textbf{F1 Macro} & \textbf{Precision} & \textbf{Recall} & \textbf{F1 Macro}  \\ \hline

\parbox[t]{2mm}{\multirow{4}{*}{\rotatebox[origin=c]{90}{\textsc{PubMed}}}} &

\textsc{SciFact} & $79.9$ & $72.6$ & $76.1$ & $73.7$ & $80.0$ & $\mathbf{76.8}$  \\
&
\textsc{PubMedQA} & $70.0$ & $70.6$ & $70.3$ & $66.7$ & $84.4$ & $\mathbf{74.5}$   \\ 
&
\textsc{HealthFC} & $62.7$ & $78.7$ & $69.7$ & $62.6$ & $84.6$ & $72.0$ \\
&
\textsc{CoVert} & $76.0$ & $83.3$ & $79.5$ & $75.6$ & $76.8$ & $76.2$  \\ \hline

\parbox[t]{2mm}{\multirow{4}{*}{\rotatebox[origin=c]{90}{\textsc{Wikipedia}}}} &

\textsc{SciFact} & $67.9$ & $83.3$ & $74.8$ & $68.8$ & $83.6$ & $75.4$ \\
&
\textsc{PubMedQA} & $65.3$ & $83.0$ & $73.1$ & $68.3$ & $78.5$ & $73.2$ \\ 
&
\textsc{HealthFC} & $62.9$ & $87.4$ & $73.1$ & $65.2$ & $92.6$ & $\mathbf{76.5}$  \\
&
\textsc{CoVert} & $72.4$ & $78.3$ & $75.2$ & $78.5$ & $86.8$ & $\mathbf{82.5}$ \\ \hline

\end{tabular}
}
\caption{\label{tab:pubmed_wikipedia} Results of final verdict prediction over four datasets using evidence retrieved from PubMed and Wikipedia.}
\end{table*}

\section{Results}

Table \ref{tab:pubmed_wikipedia} shows the performance of the claim verification system using evidence retrieved with two different techniques from two different knowledge sources, PubMed and Wikipedia. As expected, the F1 scores are lower than the oracle setting of using gold evidence from Table \ref{tab:gold}. Still, they come remarkably close to it, taking into account the complexity of finding relevant documents in a sea of 6 and 20 million articles, and further selection of relevant sentences from them, to produce a final verdict. This indicates the open-domain setting is a promising endeavor in scientific claim verification.

For both knowledge sources, the evidence from documents retrieved using semantic search outperformed the standard sparse metric BM25. Still, BM25 fares well compared to the relatively recent semantic approaches. It is also notable to observe in Table \ref{tab:pubmed_wikipedia} that BM25 excels in precision more so than recall, always beating semantic retrieval in this metric. This is not too surprising considering BM25 relies on exact keyword matching and is better suited for this use case. While BM25 slightly beats the dense BioSimCSE in precision, it is significantly outperformed in recall in the first three datasets. In deeper analysis, as we will show in the next section, we saw the dense technique would more often retrieve articles talking about the claim content using alternate naming for diseases, which led to picking up more supporting arguments for positive claims. \textsc{CoVert} is the only dataset for which BM25 performed better in the PubMed setting, in both precision and recall. Considering the noisy nature of this dataset (tweets and informal language), the inverse document frequency (idf) feature of BM25 was better at finding exact matches for important but rare keywords mentioned in the tweets and ignoring the more common but unimportant words. On the other hand, the poorer performance of the dense technique could be because the embedding was swayed in vector space due to noisy irrelevant topics from tweets.

When looking at the performance of claim verification systems over Wikipedia in Table \ref{tab:pubmed_wikipedia}, it is once again apparent that dense retrieval found more relevant documents with better evidence and outperformed the sparse retrieval. Nevertheless, in this case, the precision of BM25 was worse than BioSimCSE. In general, recall in all settings was higher than the ones from PubMed and precision lower, which shows better prediction of the positive (supported) class but also its over-prediction.

\begin{table}[htpb]
\centering
{
\def\arraystretch{1.3}
\begin{tabular}{c|ccc}
\hline
\multicolumn{1}{l}{} & \multicolumn{3}{c}{\textbf{Google Snippets}} \\ \hline
\textbf{Dataset} & \textbf{Precision} & \textbf{Recall} & \textbf{F1 Score}  \\ \hline
\textsc{SciFact} & $75.5$ & $91.5$ & $82.7$  \\
\textsc{PubMedQA} & $66.7$ & $95.6$ & $78.5$ \\ 
\textsc{HealthFC} & $62.3$ & $92.6$ & $74.5$  \\
\textsc{CoVert} & $76.4$ & $68.7$ & $72.3$  \\ \hline
\end{tabular}

}
\caption{\label{tab:google}Results of final verdict prediction over four datasets using evidence retrieved from "the whole web" (using Google).}
\end{table}

Another experiment consisted of querying "the whole web" in order to find relevant evidence for a verdict. This is a common setting explored as one of the straightforward baselines in some fact verification papers \cite{Gupta2021XFactAN, hu-etal-2022-chef} and it mimics how humans would begin the process of a claim checking. Table \ref{tab:google} reports on this performance. Considering the short nature of Google snippets, they usually do not actually provide "evidence" but commonly the verdict on the claim itself as reported on the source website containing the snippet.

\begin{table*}[!htpb]
\scriptsize
\centering
\begin{tabular}{p{22mm} p{63mm} p{63mm} }
\hline
\small \textbf{Claim }      & \small  \textbf{PubMed (semantic)  }         & \small \textbf{Wikipedia (semantic) }     \\ \hline

\textbf{Can regular intake of vitamin C prevent colds?} (\textbf{\textcolor{red}{Refuted}})                    &   Nevertheless, given the consistent effect of vitamin C on the duration and severity of colds in the regular supplementation studies, and the low cost and safety, it may be worthwhile for common cold patients to test on an individual basis whether therapeutic vitamin C is beneficial for them. \cite{Hemila2013-hc} (\textbf{\textcolor{teal}{Supported}}) & According to the most recently published \textbf{Cochrane review} on vitamin C and the common cold, one gram per day or more of vitamin C does not influence common cold incidence in the general community, i.e., it does not prevent colds.  (\href{https://en.wikipedia.org/wiki/Vitamin_C_and_the_common_cold/}{\texttt{en wiki: Vitamin C and the common cold}})  (\textbf{\textcolor{red}{Refuted}})              \\
                     &                    &                 \\
\textbf{Can lung cancer screening by computed tomography (CT) also do harm?} (\textbf{\textcolor{teal}{Supported}})                    &  Lung cancer screening with low dose computed tomography (ct) is the only method ever proven to reduce lung cancer-specific mortality in high-risk current and former cigarette smokers. We aim to explain why the risks associated with radiation exposure from lung cancer screening are very low and should not be used to avoid screening or dissuade... \cite{Frank2013-na} (\textbf{\textcolor{red}{Reftued}}) & Low-dose CT screening has been associated with falsely positive test results which may result in unneeded treatment. In a \textbf{series of studies} assessing the frequence of false positive rates, results reported that rates ranged from 8-49\%.(\href{https://en.wikipedia.org/wiki/Lung_cancer_screening}{\texttt{en wiki: Lung cancer screening}})  (\textbf{\textcolor{teal}{Supported}})              \\
                     &                    &                 \\

\textbf{Can ginkgo extract relieve the symptoms of tinnitus?} (\textbf{\textcolor{red}{Refuted}})                    &  Ginkgo biloba is a plant extract used to alleviate symptoms associated with cognitive deficits, e.g., decreased memory performance, lack of concentration, decreased alertness, tinnitus, and dizziness. Pharmacologic studies have shown that the therapeutic effect of ginkgo... \cite{Soholm1998-bw} (\textbf{\textcolor{teal}{Supported}}) & Ginkgo leaf extract is commonly used as a dietary supplement, but there is no scientific evidence that it supports human health or is effective against any disease. \textbf{Systematic reviews} have shown there is no evidence for effectiveness of ginkgo in treating high blood pressure, menopause-related cognitive decline, tinnitus, post-stroke recovery, or altitude sickness.   (\href{https://en.wikipedia.org/wiki/Ginkgo_biloba}{\texttt{en wiki: Gingko Bilboa}})  (\textbf{\textcolor{red}{Refuted}})              \\

                       \hline \\
\textbf{The most prevalent adverse events to Semaglutide are gastrointestinal.} (\textbf{\textcolor{teal}{Supported}})                    & We evaluated \underline{gastrointestinal (GI) adverse events (AEs)} with once-weekly \underline{Semaglutide} 2.4 mg in adults with overweight or obesity and their contribution to weight loss (WL).  GI AEs were more common with semaglutide 2.4 mg than placebo, but typically mild-to-moderate and transient. \cite{Wharton2022-we} (\textbf{\textcolor{teal}{Supported}}) & Possible side effects include nausea, diarrhea, vomiting, constipation, abdominal pain, headache, fatigue, indigestion/heartburn, dizziness, abdominal distension, belching, hypoglycemia (low blood glucose) in patients with type 2 diabetes, flatulence, gastroenteritis, and gastroesophageal reflux disease (GERD) (\href{https://en.wikipedia.org/wiki/Semaglutide}{\texttt{en wiki: Semaglutide}})  (\textbf{\textcolor{red}{Refuted}})              \\
                     &                    &                 \\

\textbf{Macrolides protect against myocardial infarction.} (\textbf{\textcolor{red}{Refuted}})                    & Our findings indicate that \underline{macrolide antibiotics} as a group are associated with a significant risk for \underline{MI} but not for arrhythmia and cardiovascular mortality. \cite{Gorelik2018-bm} (\textbf{\textcolor{red}{Refuted}}) & Macrolides are a class of natural products that consist of a large macrocyclic lactone ring to which one or more deoxy sugars, usually cladinose and desosamine, may be attached. (\href{https://en.wikipedia.org/wiki/Macrolide}{\texttt{en wiki: Macrolide}})  (\textbf{\textcolor{teal}{Supported}})              \\

  \hline

\end{tabular}
\caption{\label{tab:source_comparison} Example claims and retrieved evidence from the two different knowledge bases, where only one of them provided a correct final verdict.}
\end{table*}

At first glance, the performance with Google search seems impressive, especially considering that for the two most challenging datasets, \textsc{SciFact} and \textsc{PubMedQA}, the performance is improved when compared to the first two tables. A more careful look reveals this to be an artefact of data leakage and the way these two datasets were constructed (a similar phenomenon already observed in fact-checking datasets, \citet{glockner-etal-2022-missing}). Considering that in both of them the claims originate from sentences actually contained in PubMed abstracts, Google Search is powerful enough to be able to find the exact sentence that was the origin of these claims. The two other datasets, \textsc{HealthFC} and \textsc{CoVert}, give a more realistic picture of the performance of Google snippets considering they contain organic claims that originated from online users. It is interesting to see that for these two datasets Google beats both settings of PubMed but succumbs to Wikipedia as the knowledge source. This can be attributed to the fact that the simple language of claims in these two datasets can be easier to verify with Google results like blogs and news portals, as opposed to the complex language found in PubMed research publications.

\section{Discussion}

In this section, we provide further insights and a deeper look into the performance of our pipeline for open-domain claim verification of scientific claims in large knowledge sources. We do this with a qualitative analysis where we looked at what kind of documents and sentences are retrieved from different knowledge sources with different retrieval techniques and outline some common patterns with representative examples.

\begin{table*}[htpb]
\scriptsize
\centering
\begin{tabular}{p{22mm} p{63mm} p{63mm} }
\hline
\small \textbf{Claim }      & \small \textbf{BM25 (PubMed)  }         &  \small \textbf{Semantic (PubMed)  }     \\ \hline

\textbf{Do heat patches containing capsaicin help with neck pain?} (\textbf{\textcolor{red}{Refuted}})                    &  The objective of this study was to evaluate the efficacy of a hydrogel patch containing capsaicin 0.1\% compared with a placebo hydrogel patch without \textbf{capsaicin} to treat chronic myofascial \textbf{neck pain} (...) There was no significant difference between the two groups in any of the outcome measures.  (\textbf{\textcolor{red}{Refuted}})           &   In two randomized trials, a single 60-min application of the \textbf{capsaicin} 8\% patch reduced pain scores significantly more than a low-concentration (0.04\%) capsaicin control patch in patients \textbf{with PHN}.  (\textbf{\textcolor{teal}{Supported}})      \\
                     &                    &                 \\
\textbf{Does a herbal combination of rosemary, lovage, and centaury relieve symptoms of uncomplicated cystitis? } (\textbf{\textcolor{teal}{Supported}})                    &  The herbal medicinal product Canephron® N contains BNO 2103, a defined mixture of pulverized \textbf{rosemary} leaves, \textbf{centaury} herb, and \textbf{lovage} root(...) When given orally, BNO 2103 reduced inflammation and hyperalgesia in experimental cystitis in rats. (\textbf{\textcolor{teal}{Supported}}) & Rosmarinus officinalis l., rosemary, is traditionally used to treat headache and improve cardiovascular disease partly due to its vasorelaxant activity, while the vasorelaxant ingredients remain unclear. (\textbf{\textcolor{red}{Refuted}})              \\
                     &                    &                 \\
                    
\textbf{The extracellular domain of TMEM27 is cleaved in human beta cells.} (\textbf{\textcolor{teal}{Supported}})                    & Here, we report the identification and characterization of transmembrane protein 27 (\textbf{TMEM27}, collectrin) in pancreatic beta cells. (\textbf{\textcolor{teal}{Supported}}) & We also show that \textbf{TMEM2} is strongly expressed in endothelial cells in the subcapsular sinus of lymph nodes and in the liver sinusoid, two primary sites implicated in systemic HA turnover. (\textbf{\textcolor{red}{Refuted}})              \\

                      \hline \\
\textbf{Normal expression of RUNX1 causes tumorsupressing effects.} (\textbf{\textcolor{teal}{Supported}})                    &  RUNX1 is a well characterized transcription factor essential for hematopoietic differentiation and RUNX1 mutations are the cause of leukemias. runx1 is highly expressed in normal epithelium of most glands and recently has been associated with solid tumors. (\textbf{\textcolor{red}{Refuted}})   & RUNX gene over-expression inhibits growth of primary cells but transforms cells with \textbf{tumor suppressor defects}, consistent with reported associations with tumor progression. (\textbf{\textcolor{teal}{Supported}})            \\
  \hline
\end{tabular}
\caption{\label{tab:ir_comparison} Example claims and retrieved evidence from PubMed, using the two different retrieval techniques, where only one of them provided a correct final verdict.}
\end{table*}

\subsection{Popular and Specialized Claims}
The performance of Wikipedia and PubMed as a knowledge source is considerably close to each other when looking at Table \ref{tab:pubmed_wikipedia}. Nonetheless, there are differences with respect to the claim's domain and popularity.  It is evident from the tables that Wikipedia slightly outperformed PubMed for HealthFC, the dataset dealing with everyday consumer health questions, and CoVert, with social media claims related to the COVID-19 pandemic. The simple language in which these claims are posed (e.g., \textit{Does regular consumption of coffee increase the risk of heart disease such as heart attack or stroke?} as opposed to \textit{Omnivores produce less trimethylamine N-oxide from dietary I-carnitine than vegetarians}) corresponds to the more user-friendly language of Wikipedia, when compared to the often highly technical language of medical research found at PubMed.

Other than the simpler language of claims, another factor for using Wikipedia as a knowledge source is the claim's popularity and established research on it. Most claims in HealthFC are common health concerns people search for on the Internet. This means there is often systematic reviews done on these claims and Wikipedia encourages citing systematic reviews in its articles when available. We noticed our system often retrieved sentences mentioning reviews. Table \ref{tab:source_comparison} shows in first three rows how this led to the correct verdict prediction for Wikipedia, but incorrect for PubMed, since the PubMed retriever found standalone studies that might disagree from the research consensus. For 327 claims in HealthFC, combined evidence retrieved from Wikipedia mentions "systematic review" 89 times, while "Cochrane review"\footnote{Cochrane is an international organization formed to synthesize medical research findings.} is mentioned 60 times (for 1000 claims in PubMedQA, the number is 29 and 11). On the other hand, row 4 of Table \ref{tab:source_comparison} shows an example of evidence from Wikipedia being too broad and generalized, while row 5 shows a claim for which there was simply no relevant evidence in the Wikipedia article. For specialized claims concerning deeper medical knowledge or specific research hypotheses, PubMed is a superior knowledge base.
 

\subsection{Precision and Coverage}
The comparison between the two retrieval techniques in Table \ref{tab:pubmed_wikipedia} shows that semantic search outperforms BM25 in all cases, except for CoVERT on PubMed (F1: 79.5). Considering that most systems from existing work on automated fact-checking use only BM25 in their pipelines, these results can motivate future research towards deploying semantic search with different sentence embedding models. Dense retrieval's ability to deal with synonyms and paraphrases is especially important in the medical field where numerous diseases, drugs, chemical compounds can have multiple names and symbols.

While semantic search provides higher coverage, BM25 offers better precision. Table \ref{tab:pubmed_wikipedia} shows that for PubMed, using BM25 as a retrieval technique achieves higher precision for all datasets, with an especially high improvement for SciFact. The exact match of words posed in the query helps retrieving studies and documents that deal with concepts mentioned in the claim. When looking at Table \ref{tab:ir_comparison}, the first three rows show examples of claims for which dense retrieval got swayed into similar but irrelevant documents, while BM25 managed to uncover the correct ones. In the first row, capsaicin is mentioned in both, but only the one from BM25 is about neck pain. In the second row, the exact drug with specified ingredients is uncovered by BM25, while semantic search did not retrieve it. The third row shows an example of when an exact match can be important (TMEM27 vs. TMEM2). On the other hand, the fourth row shows an example of a common use case where semantic matching is beneficial -- for this claim to be matched with BM25, "tumorsuppressing" and "effects" would have to be mentioned, but dense retrieval can catch paraphrases like "tumor suppressor defects".

\subsection{Future Directions}
Based on our findings and discussion, we see the future work could focus on these direction:
\begin{itemize}
    \item  \textbf{Modeling disagreement.} We observed how different studies and sources can come to differing conclusions regarding a claim. In this paper, we chose the majority vote among the top 10 documents as the final decision, but this diminishes the information about the prediction uncertainty. This is part of the broader ML problem of learning with disagreements \cite{leonardelli-etal-2023-semeval}. The end users of fact-checking systems could appreciate the added interpretability of seeing the level of disagreement among different sources.
    \item \textbf{Assessing evidence quality.} When it comes to medical research articles from PubMed, not all of them hold the same weight, considering the research relevance. While it is hard to assess their validity of results automatically, modeling metadata aspects could give a hint on how to differently weight certain publications. Parameters such as the number of citations, the impact score and reputation of the source journal, the institutions of the authors could lead to more trustworthy results.
    Similarly, the sources of Wikipedia articles and Google search results contain website of differing reputation and credibility -- filtering to trusted domains such as university websites or academic publishers could enhance the level of trust and performance \cite{kotonya-toni-2020-explainable-automated}. Lastly, temporal aspect (date of publication) is very important since research on certain topics advances and changes with time.
     \item \textbf{Retrieval-augmented generation (RAG) for verifying claims.} Modern generative large language models (LLMs) have shown the power to both exhibit reasoning capabilities and generate coherent text for users. They already possess learned medical knowledge in their internal weights, but are prone to hallucinations. Therefore, a promising research avenue is to amplify LLMs by passing the retrieved evidence passages from sources like Wikipedia and PubMed to them \cite{pan-etal-2023-fact}. How to effectively combine this, while balancing the trade-off of readability and factuality, is an open challenge.
\end{itemize}

\section{Related Work}
\subsection{Automated Fact-Checking}

The task of automated fact-checking refers to verifying the truthfulness of a given claim using background knowledge and relevant evidence \cite{guo2022survey}. It is still mostly done manually by dedicated experts, but ongoing research efforts try to automate parts of it with NLP methods. For this purpose, many datasets have been constructed. They contain either synthetic claims generated from Wikipedia \cite{thorne-etal-2018-fever, schuster-etal-2021-get} or real-world claims found on portals dedicated to fact-checking of trending political and societal claims \cite{augenstein-etal-2019-multifc} or appearing in social media \cite{nielsen2022mumin}. Scientific fact-checking is a variation of the task that is concerned with assessing claims rooted in scientific knowledge \cite{vladika-matthes-2023-scientific}. The most prominent domains are health \cite{sarrouti2021evidence} and climate science \cite{Diggelmann2020CLIMATEFEVERAD}. 

\subsection{Open-domain Claim Verification}
Claim verification is similar to the NLP task of question answering, where the goal is to either retrieve or generate an answer to a question based on discovered evidence \cite{rogers2023qa}, and it can also be analyzed in a closed domain or open domain. In the closed domain, the evidence comes from an already provided source document. This setting is also called Machine Reading Comprehension (MRC) since the goal is to build models that are efficient in recognizing which parts of text correspond to a given query \cite{baradaran2022survey}. In the open domain, only the final answer is known and it is the goal of a system to find appropriate evidence in a large corpus of documents or other type of resources \cite{chen-yih-2020-open}. Other related tasks include Natural Language Inference \cite{vladika-matthes-2023-sebis} and Evidence Retrieval \cite{wadhwa-etal-2023-redhot}.

There have also been efforts in open-domain scientific fact verification.  \citet{wadden-etal-2022-scifact} expand the corpus of evidence research documents for the dataset \textsc{SciFact} of biomedical claims, from the original 5k to about 500k. In such a setting, they discovered significant performance drops in F1 scores of final verdict predictions. This work analyzed only one knowledge source (biomedical abstracts) and focused on data annotation in such a setting, while our paper expands the research paper corpus even further to 20 million abstracts, and analyzes other knowledge sources and retrieval methods. In \citet{pugachev2023consumer}, the authors take consumer-health question datasets and test the predictive performance of a system using PubMed and Wikipedia. A bigger focus was put on fine-tuning the models on different datasets and testing the efficiency of built-in searche engines of the respective databases. Furthermore, \citet{stammbach2023choice} test the performance of six fact-checking datasets from different domains (including encyclopedic, political) using evidence retrieved from three different knowledge bases, while looking solely at one biomedical dataset and one retrieval technique (BM25). Also related is the work by \citet{sauchuk2022role}, which shows the clear importance of the document-retrieval component of the fact-checking pipeline on the performance of the whole system. 

To the best of our knowledge, our paper features the biggest corpora (using the entirety of available PubMed and Wikipedia dumps, with 20.6M and 6.6M articles), searches "the whole web", analyzes different retrieval techniques (BM25 and semantic), and analyzes datasets of different type and purpose: expert-geared research claims (\textsc{SciFact} and \textsc{PubMedQA}), and organic user-posed health claims (\textsc{HealthFC} and \textsc{CoVert}).

\section{Conclusion}
In this paper, we conducted a number of experiments assessing the performance of a fact-verification system in an open-domain setting. Moving away from the standard setup of a small evidence corpus, we expand the knowledge sources to two large document bases (PubMed and Wikipedia), searching the whole web via Google Search API, and experiment with two retrieval techniques (sparse and dense). We measured the verdict prediction performance over four established fact-checking datasets. Our results show that searching for evidence in the open domain provides satisfyingly high F1 performance, not far from the closed-domain setting, with a room for further improvement. We conclude that the knowledge source perform comparably, with Wikipedia being better for popular and trending claims and PubMed for technical inquiries. We demonstrate the general superiority of dense retrieval techniques, with examples of where it falls short and BM25 retrieval would be beneficial. 
We hope our research will encourage more exploration of the open-domain setting in the NLP fact-checking community and addressing real-world misinformation scenarios.

\section*{Limitations}
In this study, we performed automatic assessment of claims related to medicine and health. These are two sensitive fields where misinformation, model hallucination, and incorrect evidence retrieval can lead to harmful consequences, misinformation spread, and societal effects. The automated scientific fact-verification system described in this work is still far from being safe and consistent for adoption in the real world, due to imperfect performance and drawbacks that arise. In case such an automated fact-verification system would be deployed and produce misleading verdicts, this could decrease the trust in the potential use and development of such solutions.

In our work, for easier comparison we disregard claims annotated with \textsc{Not Enough Information} due to different definitions of this label across different datasets and also absence of it in some datasets. This is an important label in claim verification, since not all claims can be conclusively assessed for their veracity. Future work should find a way to effectively include this label into model predictions. This is especially important in the scientific domain considering the constantly evolving nature of scientific knowledge, and sometimes conflicting evidence from different research studies.

Lastly, the fact-checking pipeline used in this paper is a complex system with multiple factors -- the choice of the retrieval method, of the sentence selection model, the \textit{top k} value, the NLI model, and the prediction threshold. Some incorrect predictions could have come from, e.g., faulty entailment prediction of the NLI model or other factors that do not necessarily stem from the choice of the knowledge base. Still, we put strict attention to keeping all the factors constant and frozen, to ensure a comparable setup. We focused on reporting only those phenomena and patterns that we observed were commonly occurring, after a thorough analysis of retrieved evidence for each claim. 

\section*{Acknowledgements}
This research has been supported by the German Federal Ministry of Education and Research (BMBF) grant 01IS17049 Software Campus 2.0 (TU München). We would like to thank the anonymous reviewers for helpful feedback.

\bibliography{anthology,custom}

\appendix

\end{document}